\documentclass[conference]{IEEEtran}
\usepackage{times}

\usepackage[numbers]{natbib}
\usepackage{multicol}
\usepackage[bookmarks=true]{hyperref}
\usepackage{amsmath}
\usepackage{amssymb}
\usepackage{graphicx}
\usepackage{caption}
\usepackage{subcaption}
\usepackage{array}

\begin{document}

\title{CLIP-Motion: Learning Reward Functions for Robotic Actions Using Consecutive Observations}


\author{\authorblockN{Xuzhe Dang}
\authorblockA{Faculty of Electrical Engineering\\
Czech Technical University in Prague\\
Prague, Czech\\
Email: dangxuzh@fel.cvut.cz}
\and
\authorblockN{Stefan Edelkamp}
\authorblockA{Faculty of Electrical Engineering\\
Czech Technical University in Prague\\
Prague, Czech\\
Email: stefan.edelkamp@aic.fel.cvut.cz}
}


%

\maketitle

\begin{abstract}
This paper introduces a novel reward function design method centered on the action itself. In a manner akin to STRIPS, we define several abstract motions, which can be utilized across various robotic manipulation tasks, and decompose these tasks into ordered sequences of these motions. We match these abstract motion to each action $a$ with states $s$ before and after $a$ is applied. Once an action is identified as following an abstract motion or achieving the goal state, the corresponding reward is assigned. For scenarios where only image observations are available, we fine-tune a CLIP-based model to align features $F_{o, o'}$ from image observations $o$ and $o'$ (before and after action $a$) with the features $F_{t}$ of the corresponding abstract motion description.
\end{abstract}

\IEEEpeerreviewmaketitle

\section{Introduction}
	
    Reinforcement Learning (RL) stands out within the realm of machine learning algorithms due to its distinctive approach that involves agents interacting with their environment to discover policies that maximize expected cumulative rewards. This contrasts with supervised learning, which relies on pre-determined data-label pairs for making corrections. In RL, the feedback signal comes solely from a reward function defined within the environment, making the design of this reward function crucial. A poorly designed reward function can hinder the learning process and result in a policy that predicts undesirable actions \citep{amodei2016concrete}, highlighting the importance of careful reward function engineering in RL.

   When designing a reward function for an environment, especially for robotic manipulation tasks, the common approach is to use the summed distance among objects and targets or its variants with additional rewards. For example, reach task in Fetch \citep{plappert2018multi} uses the distance between the gripper and the goal position as the reward, while pickplace task in Metaworld \citep{yu2020meta} uses the distance among the gripper, object, and goal position with an additional reward indicating whether the object is grasped by the gripper. However, such reward function designs tend to evaluate the current state rather than the action itself. A more robust approach involves action-based reward metrics that assess the quality of actions, considering factors such as action efficiency, path optimization, and dynamic interactions. Additional, in robotic manipulation tasks, to achieve the goal state, a sequence of preconditions must be achieved first. Design the reward function only with the distance among objects and target position usually miss some preconditions. For example, before robot can pull a drawer to open, the gripper must first move to the handle and hook the handle. Define the reward function by the distance among gripper, drawer's handle and goal position will ignore the condition whether gripper hooked the handle.

   Another challenge of using distance as a reward function is that it requires access to the internal states of the environment, such as the values of positions in a coordinate system. However, such features are not always easy to access, especially for real robots. In many cases, images are used to represent the observations, making it impossible to directly use a distance-based reward function.

   Inspired by how STRIPS \citep{fikes1971strips} define actions, in this paper, we propose a new reward function design method focused on the action itself. Similar to STRIPS action definitions, we define several abstract motions, such as REACH \{Object\}, GRASP \{Object\}, and MOVE \{Object\}. Each abstract motion may need to achieve preconditions before it can be applied. For example, before a robot can grasp an object, it should be close to the object. These abstract motion definitions can be shared among different robotic manipulations, and each robotic manipulation task can be decomposed as an ordered sequence of these abstract motions. Additionally, an abstract state definition is added at the end of the sequence to describe the goal state. For example, given a drawer opening task, the motion sequence is: \{ reach drawer handle, hook drawer handle, pull the drawer, drawer is opened \}. We can compute the move direction of an object with state $s$ and next state $s'$ after an action $a$ is applied. By comparing the move direction and object-target direction and determining whether the preconditions of an abstract motion are satisfied, we can decide whether an abstract motion is applied. Once we determine that the action follows an abstract motion or that the goal state is achieved, we assign the corresponding reward to this action. When only image observations are available, we fine-tune a CLIP-based model \citep{radford2021learning} to match features $F_{o, o'}$ of image observation $o$ and next image observation $o'$ after an action $a$ is applied with the features $F_{t}$ of the corresponding abstract motion description.

   To evaluate the performance of the reward functions we proposed, we select six tasks from Metaworld \citep{yu2020meta} and train DDPG \citep{lillicrap2015continuous} agents with the default reward function, our reward function with internal state, and our reward function with image observations. Additionally, following Mahmoudieh's work (ZSRM) \citep{mahmoudieh2022zero}, we fine-tune a CLIP model \citep{radford2021learning} to match the image observation of a state with the goal description and use it as a reward function as another baseline for comparison.


\section{Related Work}
\textbf{Inverse Reinforcement Learning} In Reinforcement Learning (RL), rewards are crucial as they serve as the primary feedback mechanism to evaluate the efficacy of actions dictated by a policy. However, designing effective reward functions is challenging due to the complexity of tasks, unclear properties for informing rewards, and prevalent reward sparsity in various environments. Consequently, robust reward function design has become a central focus in RL research. To mitigate the absence of explicit reward functions, researchers have developed various approaches, such as leveraging intrinsic motivation as an additional reward\citep{pathak2017curiosity, zheng2018learning, pan2019risk}. Alternative methods are proposed include training agents to infer goal conditions from observations, transforming no-reward scenarios into ones with sparse rewards, and utilizing hindsight experience replay to reframe unsuccessful attempts as valuable learning experiences~\citep{andrychowicz2017hindsight, fang2018dher, fang2019curriculum, yang2021mher}. Expert demonstrations play a pivotal role in imitation learning, providing a rich source of guidance for agent behavior. Reward functions derived from these demonstrations offer a pathway to knowledge transfer, applicable within and across domains \citep{piot2016bridging, reddy2019sqil, yu2020intrinsic}. The integration of language with reinforcement learning opens new avenues for tasks requiring complex interactions, such as robotic manipulation and navigation, with Vision-Language Models enhancing reward function learning through the comparison of visual observations with textual goal descriptions~\citep{mahmoudieh2022zero, du2023vision}.

\textbf{Vision-Language Models} Vision-Language Models (VLMs) have become a prominent area of research due to their ability to understand and generate descriptions from visual inputs. Early attempts at integrating vision and language focused on image captioning \citep{li2017image, ding2019image, liu2020image} and visual question answering (VQA) \citep{zhou2020unified, ma2021joint, lu2023multiscale}. The advent of transformer architectures revolutionized VLMs. Transformers \citep{vaswani2017attention} were adapted for multimodal tasks, enabling more effective cross-modal attention mechanisms. Models like ViLBERT \citep{lu2019vilbert} and LXMERT \citep{tan2019lxmert} extended the BERT architecture \citep{devlin2018bert} to handle both visual and textual inputs. The trend towards large-scale pretraining has significantly impacted VLMs. Models like CLIP \citep{radford2021learning} and ALIGN \citep{jia2021scaling} leverage extensive datasets to learn robust multimodal representations. 


\section{Preliminaries}
\subsection{MDP}
A Markov Decision Process (MDP) is formally defined by a tuple $(O, A, P, R, \gamma)$, where $O$ represents a finite set of observations, $A$ denotes a finite set of actions, $P(o'|o, a)$ is the transition function that gives the probability of transitioning from observation $o$ to observation $o'$ after taking action $a$, $R(o, a)$ is the reward function specifying the immediate reward received after action $a$ is performed in observation $o$, and $\gamma \in (0, 1)$ is the discount factor. In our method, however, the reward function $R(o, o')$ specifies the immediate reward by the observation $o$ and the next observation $o'$ after action $a$ is applied. The policy $\pi(a|o)$ is a mapping from observations to actions, guiding the decision-making process.

The primary objective in an MDP is to determine an optimal policy $\pi$ that maximizes the expected discounted total reward. In a finite horizon MDP, this objective is mathematically expressed as:
\begin{equation}
    \max_{\pi} \mathbb{E}_{\tau} \left[ \sum_{t=0}^{T} \gamma^t R(o_t, o_{t+1}) \right]
\end{equation}

where $\tau$ denotes a trajectory following the policy $\pi$, such that actions $a_t$ are sampled according to $\pi(a_t|o_t)$, and observations $o_{t+1}$ follow the transition probabilities $P(o_{t+1}|o_t, a_t)$.

\subsection{CLIP}
Contrastive Language-Image Pre-training (CLIP) \citep{radford2021learning}  has effectively showcased the capability of learning open-set visual concepts. CLIP architecture comprises two encoders: one for images and one for text. The image encoder, which can be either a ResNet \citep{he2016deep} or a Vision Transformer (ViT) \citep{dosovitskiy2020image}, converts an image into a feature vector. The text encoder, a Transformer model, processes a sequence of word tokens to generate a corresponding vectorized representation.

CLIP uses a contrastive loss to learn a joint embedding space for both modalities. Give a batch of image-text pairs, CLIP maximizes the cosine similarity between each image and its corresponding text, while minimizing the cosine similarities with all other unmatched texts. Let $ x $ represent the image features generated by the image encoder, and let $\{w_i\}_{i=1}^K $ be a set of weight vectors produced by the text encoder, each representing a category (with $ K $ categories in total). Each $ w_i $ is derived from a prompt, such as “a photo of a {class},” where the “{class}” token is replaced with the $ i $-th class name. The prediction probability is given by:

\begin{equation}
p(y|x) = \frac{\exp(\text{sim}(x, w_y)/\tau)}{\sum_{i=1}^K \exp(\text{sim}(x, w_i)/\tau)}
\end{equation}

where $\text{sim}(\cdot, \cdot)$ denotes cosine similarity and $\tau$ is a learned temperature parameter.

\subsection{DDPG}
Deep Deterministic Policy Gradient (DDPG) \citep{lillicrap2015continuous} is an actor-critic algorithm designed for continuous control tasks. It simultaneously learns a Q-function $ Q_{\theta} $ and a deterministic policy $ \pi_{\phi} $. DDPG employs Q-learning \citep{watkins1992q} to learn $Q_{\theta}$ by minimizing the one-step Bellman residual:

\begin{equation}
\begin{aligned}
    J_{\theta}(\mathcal{D}) = \mathbb{E}_{(x_t,a_t,r_t,x_{t+1}) \sim \mathcal{D}} 
    \Big[ & \left( Q_{\theta}(x_t, a_t) - r_t \right. \\
    & \left. - \gamma Q_{\bar{\theta}}(x_{t+1}, \pi_{\phi}(x_{t+1})) \right)^2 \Big]
\end{aligned}
\end{equation}

The policy $ \pi_{\phi} $ is learned using the Deterministic Policy Gradient (DPG) algorithm \citep{silver2014deterministic}, which maximizes:

\begin{equation}
    J_{\phi}(\mathcal{D}) = \mathbb{E}_{x_t \sim \mathcal{D}} \left[ Q_{\theta}(x_t, \pi_{\phi}(x_t)) \right]
\end{equation}

This optimization ensures that $ \pi_{\phi}(x_t) $ approximates $ \arg\max_{a} Q_{\theta}(x_t, a)$. In this context, $ \mathcal{D} $ represents a replay buffer of environment transitions, and $ \bar{\theta} $ is an exponential moving average of the weights.


\section{Method}

\begin{figure*}[ht]
    \begin{center}
        \includegraphics[width=1.0\textwidth,trim={0 8cm 0 8cm},clip]{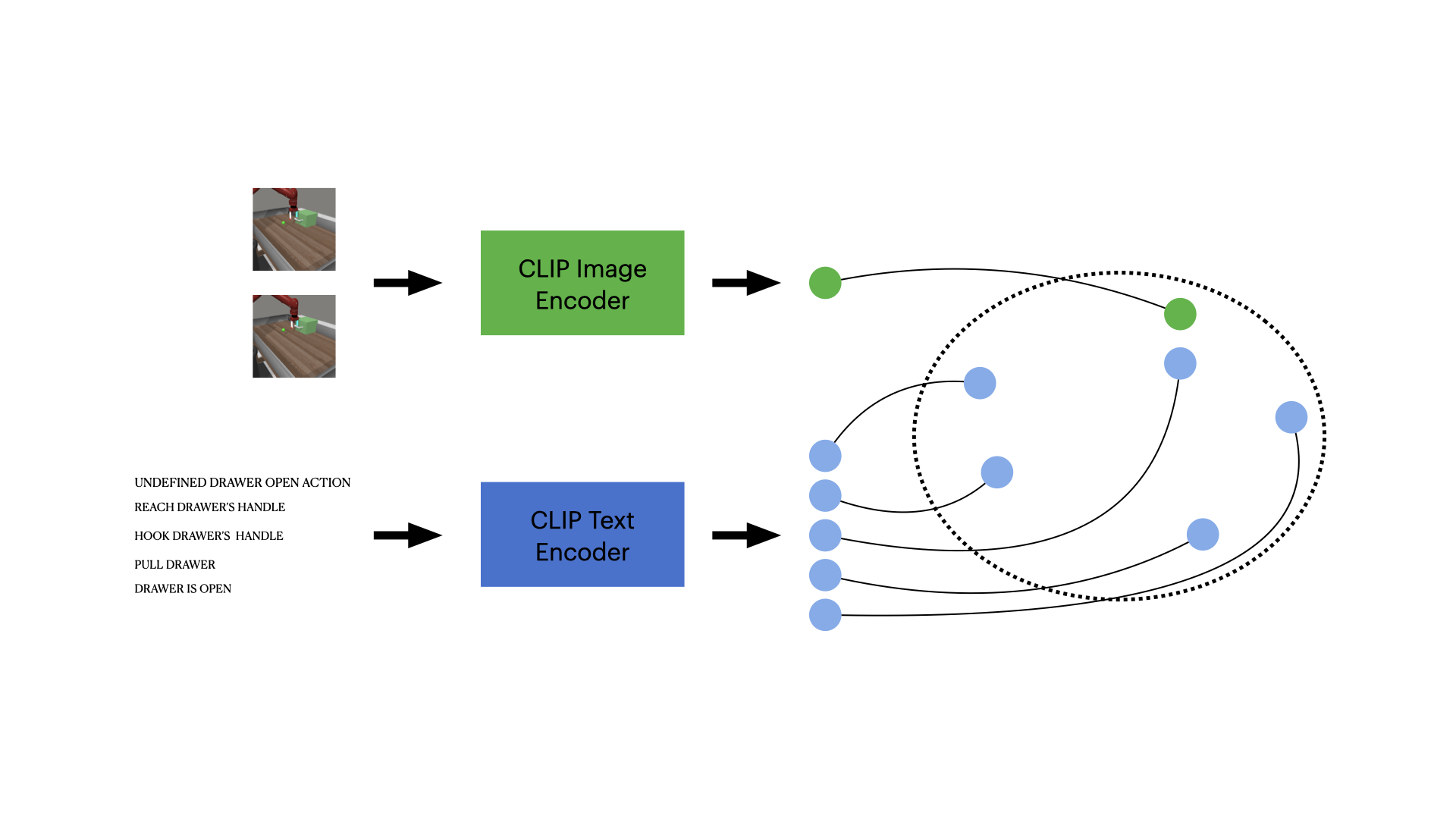}
    \end{center}
    \caption{The image observations are encoded into observation features by the CLIP Image Encoder, and the text descriptions of abstract motions are encoded into text features by the CLIP Text Encoder. Cosine similarity is computed for each pair of observation-text features, and the action is determined to match the abstract motion which has the maximum similarity.}
    \label{fig:match_imgs}
\end{figure*}

\subsection{Abstract Motion Definition}
STRIPS (Stanford Research Institute Problem Solver) planning is an approach used in the field of artificial intelligence (AI) for automated planning and scheduling. In STRIPS, actions are defined using three key components:

\textbf{Preconditions}: Conditions that must be true before an action can be executed. Preconditions describe the necessary state of the world for the action to be applicable.

\textbf{Add List}: Describes the effects of the action that add to the current state. These are the conditions that will become true after the action is performed.

\textbf{Delete List}: Describes the effects of the action that remove from the current state. These are the conditions that will become false after the action is performed.

Inspired by the action definition of STRIPS, we propose our reward function design method matching actions to abstract motions. We first define several abstract motions, which describe common actions in various robotic manipulation scenarios, including REACH \{Object\}, GRASP \{Object\}, MOVE \{Object\} TO \{Object\}, HOOK \{Object\}, PUSH \{Object\} TO \{Object\}, PULL \{Object\} TO \{Object\}, and SLIDE \{Object\} TO \{Object\}. These abstract motions may have some preconditions that must be achieved before they can be applied. Additionally, for each task, we define its goal state description. Once the abstract motions and goal state are defined, we manually decompose each robotic manipulation task into an ordered sequence of these abstract motions, appending the goal state to the end. When an action $a$ is applied, we use state $s$ and next state $s$ to determine whether the action $a$ matches the abstract motions in the ordered sequence. For example, we can determine whether the gripper is reaching the target location by comparing the cosine similarity between the gripper's move direction and the gripper-target direction, or whether the gripper is grasping an object by computing whether the gripper is closing when it is near to the object. For the actions don't fit into any abstract motions, we define as an unwanted motions for each tasks, with reward set to 0.

\begin{figure*}[ht]
    \begin{center}
        \includegraphics[width=1.0\textwidth,trim={0 8cm 0 8cm},clip]{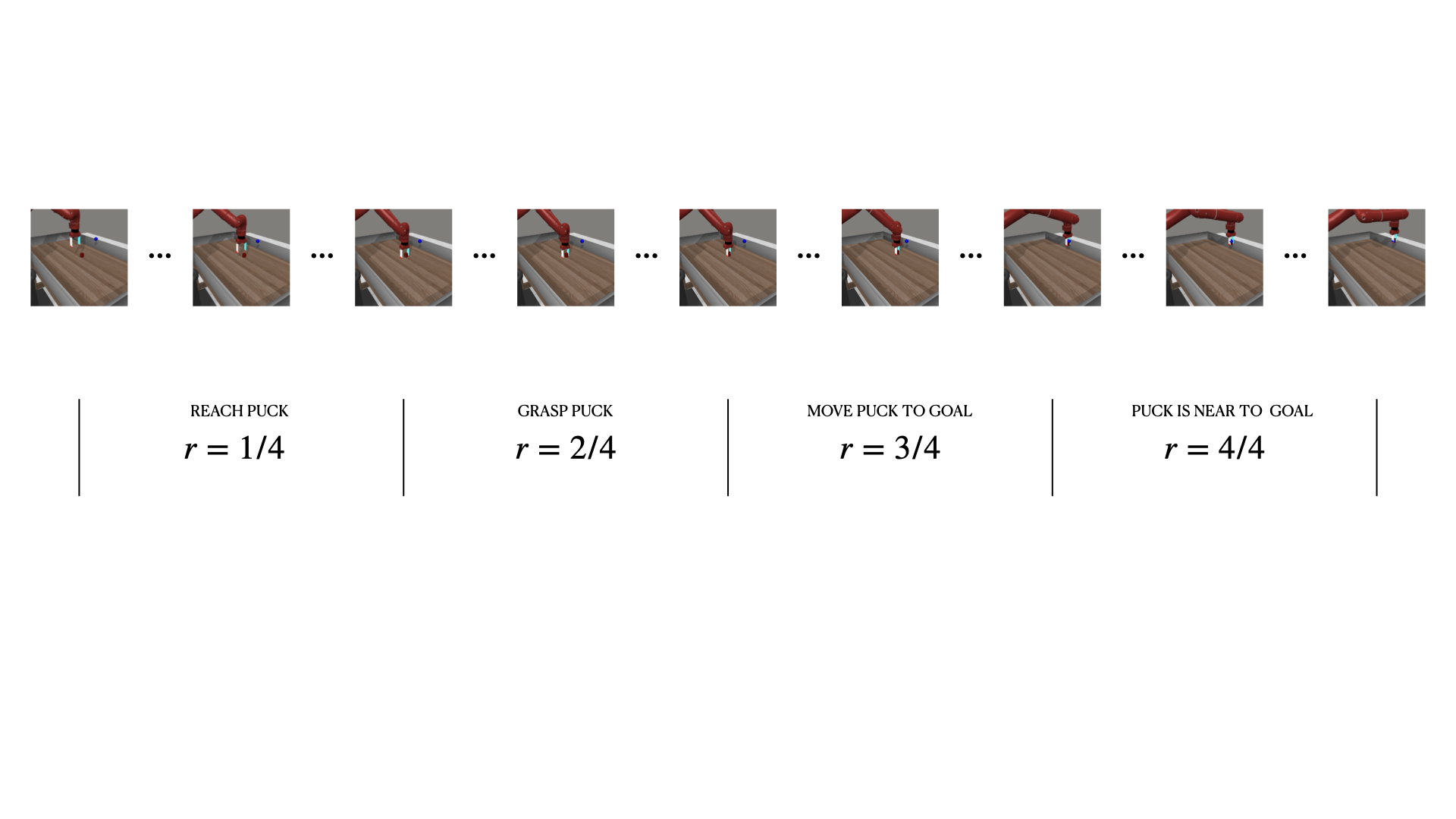}
    \end{center}
    \caption{A "pick place" task can be decomposed into an abstract motion sequence: reach puck, grasp puck, move puck to goal, and puck is near the goal. Actions determined to match each abstract motion are assigned corresponding rewards.}
    \label{fig:reward_assign}
\end{figure*}

\subsection{Fine Tuning CLIP Model}
When the internal state can't be accessed, we extend our reward function to visual observations by applying the CLIP model with ResNet 50 to match the images of two states and the text descriptions of abstract motions.

The original CLIP model is tailored for pairing single images with textual descriptions. However, our approach involves utilizing two observations as input images, introducing a notable challenge. This divergence from the standard three-channel RGB format, expected by most pretrained image models, necessitates a novel solution. We implemented a convolutional channel expansion technique~\cite{wang2022vrl3} to modify the model's input layer, allowing it to handle the augmented observation channels. This adjustment ensures that the model remains adept at analyzing both temporal and spatial dimensions of the input data.

Convolutional channel expansion is achieved by replicating the ResNet model's weight matrix by a factor of $m$ and then adjusting these weights to ${1}/{m}$
of their initial values, here, in our method, $m=2$. This strategy allows for the simultaneous interpretation of temporal and spatial data characteristics, greatly augmenting the model's aptitude for navigating dynamic settings. An essential advantage of this technique is that it boosts model performance while retaining the pretrained network's weights unchanged, thus upholding its operational integrity and efficiency.

To fine-tune the CLIP model, we collect 100,000 samples for each task. Each sample contains the image observations before and after an action is applied, and the abstract motions of the action. Because the CLIP model is not only learning to determine whether actions that fit into abstract motions but also to identify actions of undefined motions, we need to collect samples from unsuccessful episodes as well. Thus, we collect our data not only from well-trained policies but also from policies that are not well-trained and from a uniform sample policy.

Once the CLIP model is fine-tuned, we can encode the image observations $o$ and $o'$ into $F_{o, o'}$ using the CLIP Image Encoder. The text descriptions of abstract motions $t_{m}$ are encoded into $F_{t}$. We assign the abstract motion that has the maximum cosine similarity between $F_{o, o'}$ and $F_{t}^{i}$ to the action $a$, as shown in Figure \ref{fig:match_imgs}.

\subsection{Reward Assignment}
Once the abstract motion of an action $a$ is determined, we need to assign the reward value to this action $a$. Because the task is decomposed to ordered sequence of abstract motions, the reward should be incremental to encourage the policy learns to perform actions following the order. In our study, we simply assigning the reward value for each abstract motions of a task by:
\begin{equation}
    r = \frac{I}{n}
\end{equation}
where r represents the reward value, I represents the index of abstraction motion in sequence and n represents the number of abstract motions in tasks. An example is shown in Figure \ref{fig:reward_assign}.


\begin{figure*}[tbp]
    \centering
    \begin{subfigure}[b]{0.45\textwidth}
        \centering
        \includegraphics[width=\textwidth]{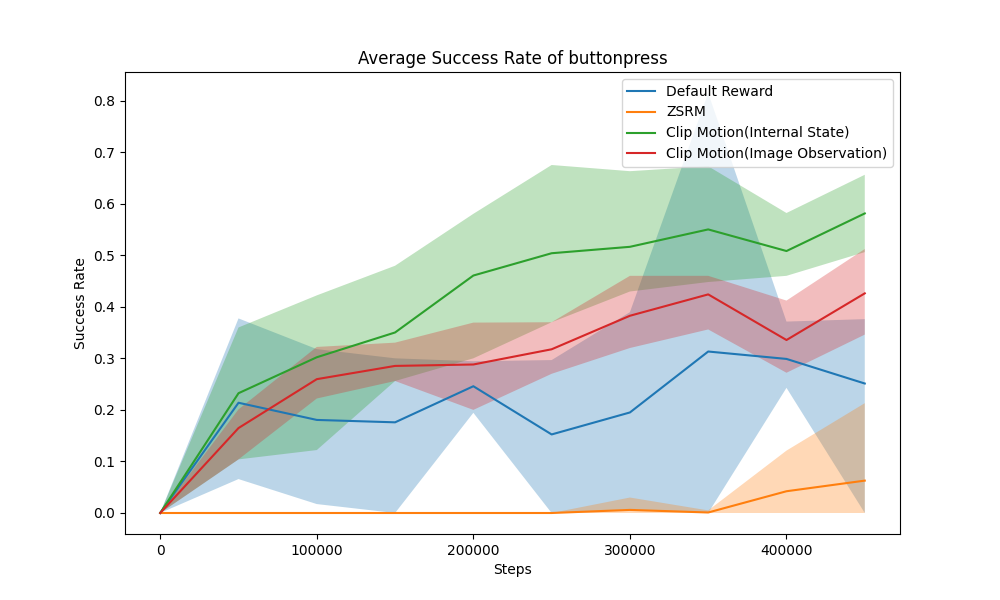}
        
        \caption{Button Press Task Success Rate}
        \label{fig:sub1}
    \end{subfigure}
    \hfill
    \begin{subfigure}[b]{0.45\textwidth}
        \centering
        \includegraphics[width=\textwidth]{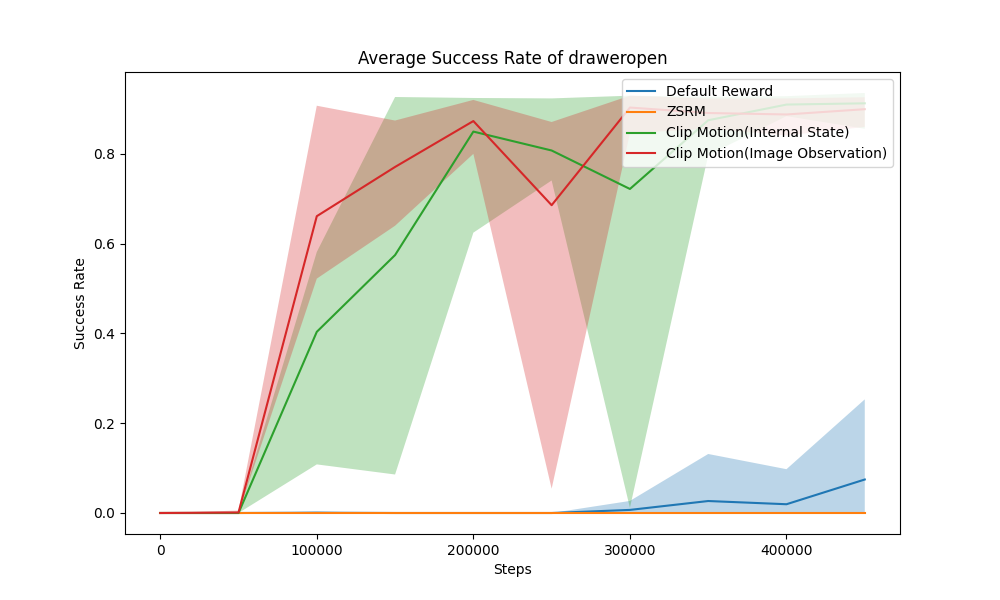}
        \caption{Drawer Open Task Success Rate}
        \label{fig:sub2}
    \end{subfigure}
    \hfill
    \begin{subfigure}[b]{0.45\textwidth}
        \centering
        \includegraphics[width=\textwidth]{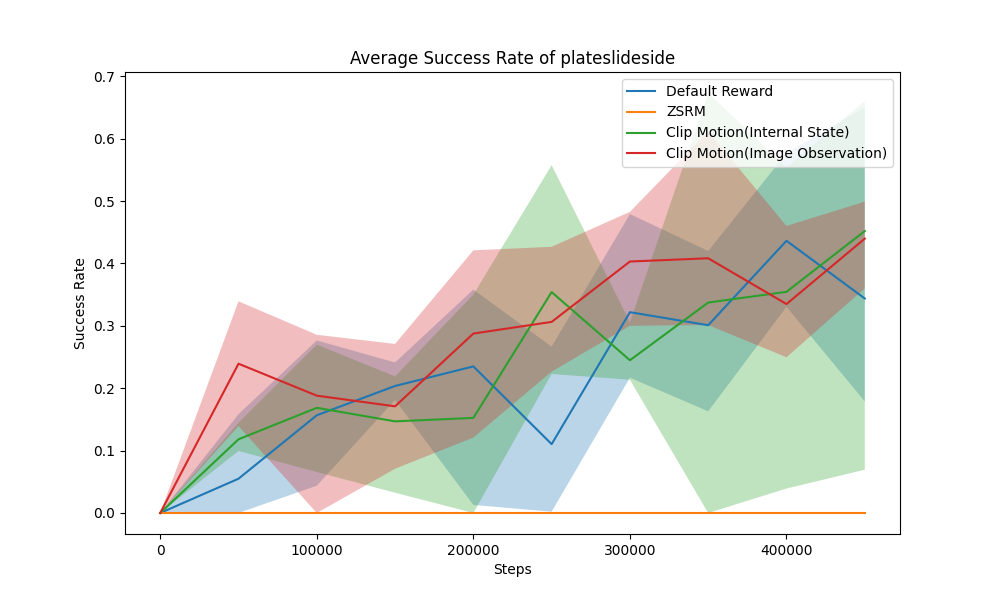}
        \caption{Plate Slide Side Task Success Rate}
        \label{fig:sub2}
    \end{subfigure}
    \hfill
    \begin{subfigure}[b]{0.45\textwidth}
        \centering
        \includegraphics[width=\textwidth]{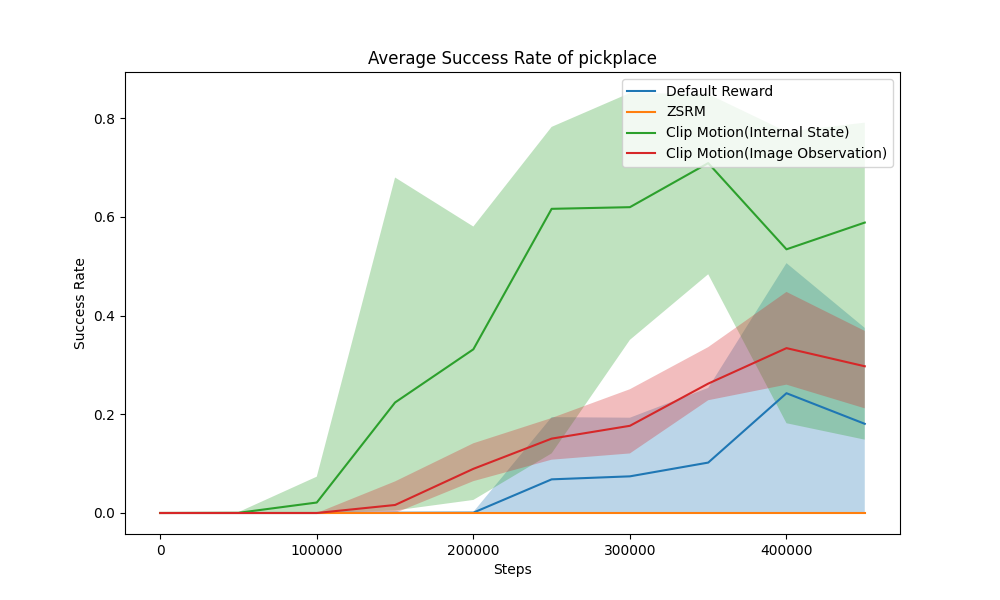}
        \caption{Pick Place Task Success Rate}
        \label{fig:sub2}
    \end{subfigure}
    \hfill
    \begin{subfigure}[b]{0.45\textwidth}
        \centering
        \includegraphics[width=\textwidth]{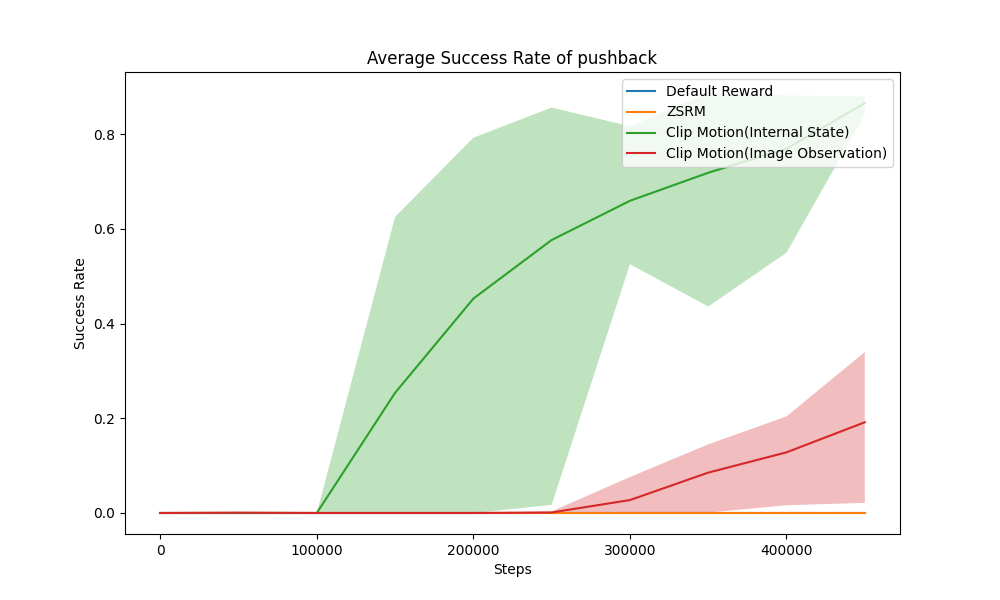}
        \caption{Push Back Task Success Rate}
        \label{fig:sub2}
    \end{subfigure}
    \hfill
    \begin{subfigure}[b]{0.45\textwidth}
        \centering
        \includegraphics[width=\textwidth]{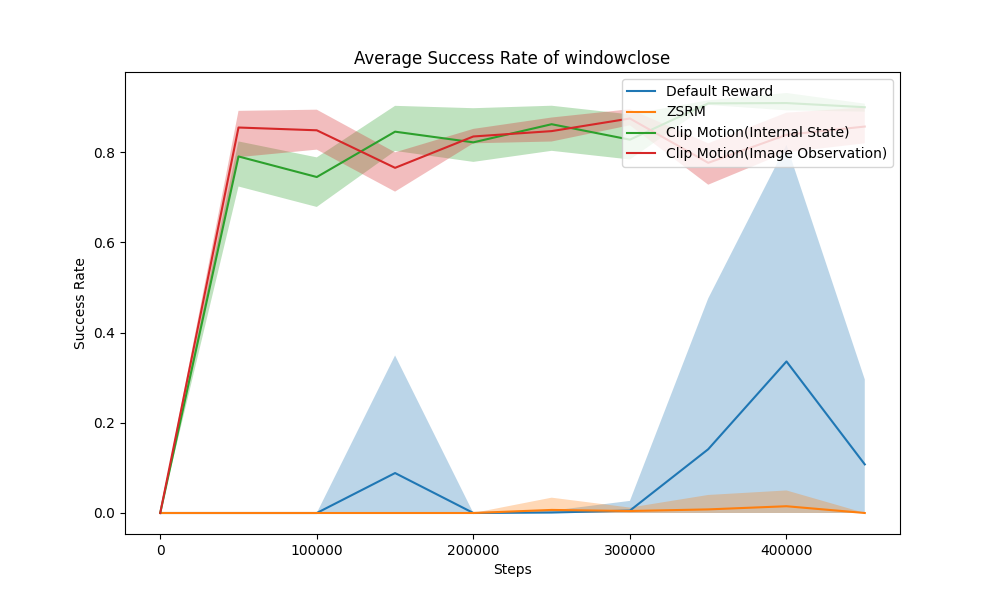 }
        \caption{Window Close Task Success Rate}
        \label{fig:sub2}
    \end{subfigure}
    \caption{Success rate in different Metaworld environments.}
    \label{fig:results}
\end{figure*}

\section{Experiment}

\subsection{Implementation Detail}
Our training method employs the Deep Deterministic Policy Gradient (DDPG) algorithm \citep{lillicrap2015continuous} as the backbone reinforcement learning framework, ensuring effective learning in continuous action spaces. Similar to DrQv2 \citep{yarats2021mastering}, in our study, we used a shared encoder to process the observations for both actor and critic. We apply layer normalization \citep{ba2016layer} to each MLP layer within the encoder, actor and critic. During our experiments, we observed that layer normalization contributes significantly to the stability and acceleration of the learning process. Additionally, we replaced the ReLU activation function \citep{agarap2018deep} with SiLU \citep{elfwing2018sigmoid}, which further enhances the model's performance. To mitigate overestimation bias in the target value, we incorporate clipped double Q-learning \citep{van2016deep}, training two separate Q-networks $Q_{\theta_{1}}$ and $Q_{\theta_{2}}$. The loss of the Q-networks is computed by:
\begin{equation}
    L_\theta^k (\mathcal{D}) = \mathbb{E}_{\tau \sim \mathcal{D}} \left[ (Q_\theta^k (Z_{t}, a_{t}) - y)^2 \right] \quad \forall k \in \{1, 2\}
\end{equation}
with the TD target $y$:
\begin{equation}
    y =  \gamma r_{t+i} + \gamma \min_{k=1,2} Q_{\bar{\theta}^k} (Z_{t}, a_{t})
\end{equation}
During the training process, the encoder is updated solely in conjunction with the critic networks. This means the encoder's parameters are adjusted based on the gradients derived from the Q-networks' loss function.

To stable the learning, we added an action regularization to the actor loss as:
\begin{equation}
    L_{\phi}(\mathcal{D}) = -\mathbb{E}_{\tau \sim \mathcal{D}} \left[ \min_{k=1,2} Q_{\theta_k}(h_t, a_t) \right] + \lambda \cdot \frac{1}{N} \sum_{i=1}^{N} \| a_i \|_2^2
\end{equation}
, where $\lambda$ is set to $1e^{-3}$.

To accelerate the training process, we employ RAY \citep{moritz2018ray} to collect samples in parallel. Additionally, we reset the task scenario in Metaworld \citep{yu2020meta} after 100 steps instead of the default 500 steps during each episode. In our study, we find that the diversity of scenarios accelerates learning, and shorter episodes allow us to gather samples from more scenarios within the same number of samples. Instead of updating parameters during the sample collection phase, we perform multiple update steps only after the collection of an episode is completed. This approach ensures efficient use of computational resources and improves the overall training speed.

\subsection{Results}
 The effectiveness of our methodology was evaluated on a suite of six varied tasks, as delineated in \textit{Metaworldv2} \citep{yu2020meta}. Training for each task was iteratively conducted using three distinct random seeds to ensure generalizability and consistency of our findings.

 To evaluate our approach's performance comprehensively, we trained each task under various reward models, including:

\begin{itemize} \setlength\itemsep{0em}
    \item \textbf{Default Reward:} This model, the Metaworld's default, calculates rewards by summing the distances between the gripper, objects, and goal positions, incorporating the gripper finger state for additional rewards.
    \item \textbf{ZSRM (Zero-Shot Reward Modeling) \citep{mahmoudieh2022zero}:} ZSRM leverages a CLIP model to ascertain if a visual observation meets the goal achievement conditions, as described linguistically. We fine-tuned a pretrained CLIP model using a dataset curated for our study.
    \item \textbf{Clip-Motion (Internal State):} When internal states are accessible, we compute the movement vectors for a specified object across two consecutive observations to determine whether the action adheres to abstract motions, assigning rewards accordingly.
    \item \textbf{CLIP-Motion (Image Observation):} This model processes visual observations and textual descriptions of abstract motions. We assign a predefined reward value for each described abstract motion.
\end{itemize}

The outcomes of our experiments are illustrated in Figure~\ref{fig:results}, where we use the success rates for each task as metrics to evaluate the performance of agents. The agents are trained within 500K samples and evaluated every 50K samples. As shown in the results, the agents trained with our reward function using internal state not only achieve the highest success rate after 500K samples but also learn to solve the tasks more quickly in all tasks. Moreover, in tasks like drawer open and window close, the policies trained with our reward function are more stable. When only image observations are used to compute the reward function, our reward function with image inputs outperforms both the default reward function and ZRSM. However, due to prediction errors, the performance of the policy with the image observation reward function is sometimes much worse than the policy with the internal state reward function.


\section{Limitation}
Our current work exhibits some limitations. Firstly, the manual construction of sequence of abstract motions, as practiced in this study, presents a potential challenge when dealing with highly intricate tasks. As tasks grow in complexity, the manual approach becomes less feasible. Addressing this concern requires an exploration of strategies to automate the decomposition of complex tasks into sequences of fundamental motions. To tackle this challenge, we are actively exploring methodologies such as Large Language Models (LLM)~\citep{touvron2023llama, achiam2023gpt, touvron2023llama, jiang2023mistral} and Hierarchical Reinforcement Learning~\citep{shu2017hierarchical, sohn2018hierarchical, takanobu2019hierarchical, florensa2018automatic}, which have the potential to streamline task decomposition.

Secondly, the reliance on the collection of robotic data remains a significant component of our ongoing work. As part of our future initiatives, we are enthusiastic about harnessing the potential of transferring knowledge from human demonstrations. This strategic approach offers the promising prospect of diminishing the necessity for extensive robotic data collection, potentially expediting the learning process and boosting overall efficiency.


\section{Conclusion}
In this paper, we propose a new reward function design method focused on the action itself. We define several abstract motions which are applicable across different robotic manipulations, allowing each task to be decomposed into an ordered sequence of these motions, with an abstract state definition added at the end to describe the goal state.

In our approach, we determine whether an action follows an abstract motion with state $s$ and next state $s'$. Once confirmed, we assign the corresponding reward to the action. For scenarios where only image observations are available, we fine-tune a CLIP-based model \citep{radford2021learning} to match the features of image observations before and after an action with the features of the corresponding abstract motion description.

Our experiments demonstrate that agents trained with our reward function using internal state information achieve the highest success rates after 500K samples and learn to solve tasks more quickly. Additionally, policies trained with our reward function exhibit greater stability in tasks such as drawer open and window close. When using image observations to compute the reward function, our method outperforms both the default reward function and ZRSM. However, due to prediction errors, the performance of the policy with the image observation reward function can be inferior to the policy with the internal state reward function in some tasks.

Overall, our proposed reward function design method shows significant promise in improving the efficiency and stability of training robotic manipulation tasks, particularly when internal state information is accessible.

\section{Acknowledgements}
The presented work has been supported by the Czech Science Foundation (GA\v{C}R) under the research project number 22-30043S.


\bibliographystyle{plainnat}
\bibliography{references}

\end{document}